\PassOptionsToPackage{table}{xcolor}

\documentclass{article} 
\usepackage{iclr2025_conference,times}

\usepackage{amsmath,amsfonts,bm}

\def\eqref#1{equation~\ref{#1}}

\def\1{\bm{1}}

\DeclareMathAlphabet{\mathsfit}{\encodingdefault}{\sfdefault}{m}{sl}
\SetMathAlphabet{\mathsfit}{bold}{\encodingdefault}{\sfdefault}{bx}{n}

\usepackage{enumitem}
\usepackage{url}
\definecolor{citegreen}{RGB}{2, 142, 2}
\usepackage[pagebackref=false,breaklinks=true,colorlinks=true,citecolor=citegreen,bookmarks=false]{hyperref}
\usepackage{amsthm}
\usepackage{thmtools, thm-restate}
\usepackage{amsmath}
\usepackage{amsfonts}
\usepackage{amssymb}
\usepackage{mathtools}
\usepackage{float}
\usepackage{subfig}
\usepackage{multicol}
\usepackage{multirow}
\usepackage{booktabs}
\usepackage{wrapfig}
\usepackage{graphicx}
\usepackage{subcaption}
\usepackage{graphicx}

\definecolor{rowblue}{RGB}{230,245,255}
\definecolor{podiblue}{rgb}{0.9, 0.92, 1.0}
\definecolor{deepred}{RGB}{170,0,0}

\newtheorem*{rep@proposition}{Proposition \ref{prop:better}}
\newenvironment{proposition@repeat}[1]{
  \def\thetheorem{\ref{#1}}
  \rep@proposition
}{\endrep@proposition}

\title{UniFork: Exploring Modality Alignment for Unified Multimodal Understanding and Generation}
\iclrfinalcopy

\author{Teng Li$^{1,2}$, Quanfeng Lu$^{3,2}$, Lirui Zhao$^{2}$, Hao Li$^{2}$, \\
\textbf{Xizhou Zhu$^2$}, \textbf{Yu Qiao$^2$}, \textbf{Jun Zhang\thanks{Corresponding Authors.}~~$^{1}$}, \textbf{Wenqi Shao$^{*2}$} \\
\\[0.3em]
$^{1}$HKUST,
~~$^2$Shanghai AI Laboratory, ~~$^3$SJTU \\
}

\begin{document}

\maketitle

\begin{abstract}
Unified image understanding and generation has emerged as a promising paradigm in multimodal artificial intelligence. Despite recent progress, the optimal architectural design for such unified models remains an open challenge. In this work, we start by analyzing the modality alignment behaviors of task-specific expert models for understanding and generation, as well as current unified models. Our analysis reveals a crucial observation: understanding tasks benefit from a progressively increasing modality alignment across network depth, which helps build up semantic information for better comprehension; In contrast, generation tasks follow a different trend—modality alignment increases in the early layers but decreases in the deep layers to recover spatial details. These divergent alignment patterns create a fundamental conflict in fully shared Transformer backbones, where a uniform representational flow often leads to performance compromises across two tasks. Motivated by this finding, we introduce \textbf{UniFork}, a novel Y-shaped architecture that shares the shallow layers for cross-task representation learning, while employing task-specific branches in deeper layers to avoid task interference. This design effectively balances shared learning and task specialization. Through extensive ablation experiments, we demonstrate that Unifork consistently outperforms conventional fully shared Transformer architectures, and achieves performance on par with or better than task-specific models. Our code is available at {\small \url{https://github.com/tliby/UniFork}}.

\end{abstract}

\section{Introduction}

Recent works~\citep{showo, li2025synergen, bagel, zhang2025survey} have demonstrated significant progress in unified multimodal generation and understanding. By projecting both language and vision signals into a shared embedding space and arranging them in various ways, it becomes feasible to perform both image understanding and generation within a single Transformer architecture. However, despite sharing such a common paradigm, the objectives of generation and understanding tasks are inherently different~\citep{wu2025janus, januspro}. Image generation emphasizes the fidelity and aesthetic quality of visual outputs, focusing on pixel-level details such as texture and color. In contrast, image understanding centers on high-level semantic comprehension, such as identifying objects, interpreting spatial relationships, and reasoning about scene content. This fundamental divergence makes it notoriously challenging to unify the two tasks .

To address the task discrepancy issue, some recent approaches~\citep{wu2025janus, januspro} adopt distinct semantic and spatial image representations tailored to understanding and generation respectively. Other methods introduce diffusion optimization objectives~\citep{showo, zhou2024transfusion} or external models~\citep{ge2024seedx, ming-uni} to decode spatial features for image generation. Although these designs can enhance task-specific performance, they often undermine the simplicity and elegance of the original next-token prediction (NTP) paradigm in large language models (LLM). In addition, during supervised fine-tuning (SFT), meticulous data balancing is typically required to maintain performance across tasks. Furthermore, the intrinsic relationship between generation and understanding remains largely unexplored, raising important questions about how these tasks might complement each other within a unified framework.

In this work, we investigate the relationship between image understanding and generation through the lens of feature alignment between image and language tokens. We find that these two tasks exhibit distinct alignment patterns: image understanding benefits from progressively increasing alignment across network depth to build semantic representations, whereas image generation relies on strong early-layer alignment followed by weakened coupling in later layers to enable fine-grained visual synthesis. Moreover, employing a fully shared Transformer backbone under the NTP modeling paradigm enforces a representational compromise between the two tasks. These findings underscore the importance of accounting for the divergent alignment patterns of understanding and generation when designing unified models, in order to achieve optimal performance across both tasks.


\begin{figure}[!t]
    \centering
    \includegraphics[width=\linewidth]{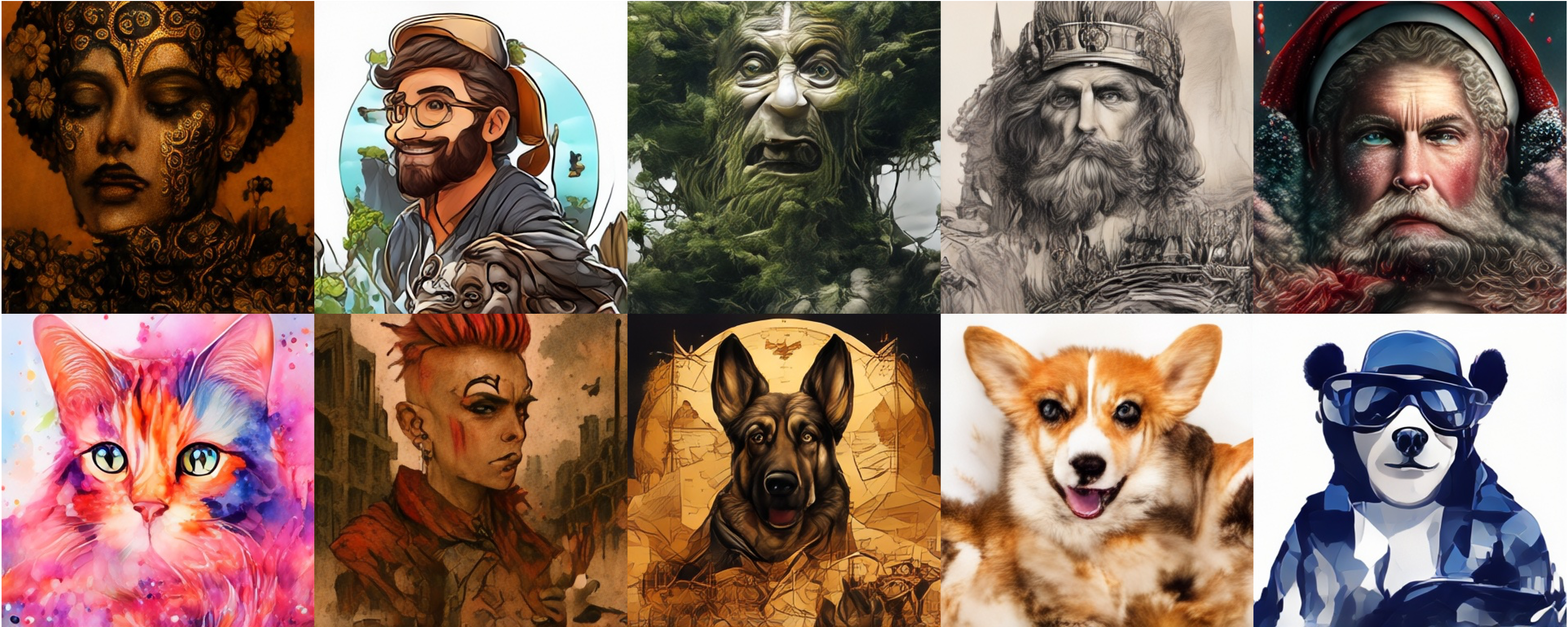}
    \caption{Text-to-image generation results by UniFork in 384×384 resolution.}
    \label{fig:1}
\end{figure}

Building upon the observation, we propose \textbf{UniFork}, a Y-shaped architecture for unified image understanding and generation. Specifically, the early layers of the Transformer backbone are shared across both tasks to enable cross-task semantic learning. In the latter layers, we introduce task-specific branches—two structurally identical yet independently parameterized modules. The understanding branch refines semantic representations, whereas the generation branch reconstructs spatial details. By decoupling the task-specific representation learning in the later layers, UniFork effectively alleviates the representational conflict arising from divergent alignment patterns. An additional advantage of UniFork lies in its training flexibility. During the final SFT stage, task-specific parameters can be independently optimized using their respective datasets, eliminating the need for delicate data ratio adjust. To validate the effectiveness of our design, we conduct extensive ablation studies showing that UniFork outperforms fully shared architectures and achieves performance comparable to task-specific expert models. Furthermore, with moderate scaling based on Qwen2.5-0.5B LLM~\citep{qwen2.5}, UniFork outperforms the state-of-the-art unified models trained at similar scale. Our main contributions are summarized as follows:
\begin{itemize}[leftmargin=*]
    \item We analyze task-specific modality alignment patterns in expert models, highlighting the differing needs of image understanding and generation, and providing insights for unified model design.
    \item We propose UniFork, a Y-shaped architecture that decouples task-specific learning in the later layers while retaining shared semantic representation learning in the early layers. This design enables effective cross-task learning and alleviates performance conflicts between tasks.
    \item Comprehensive ablation studies demonstrate that UniFork outperforms fully shared Transformer architectures. With moderate scaling, our approach achieves significantly improved performance on both understanding and generation tasks.

\end{itemize}

\section{Related Work}

\begin{figure}[!t]
    \centering
    \includegraphics[width=\linewidth]{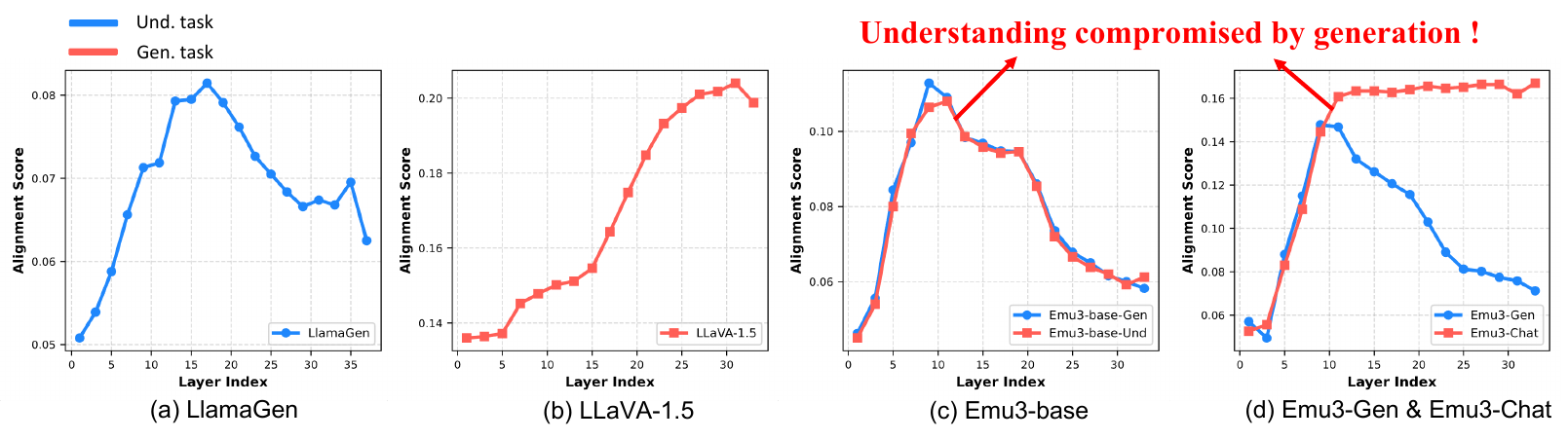}
\caption{\textbf{Modality alignment analysis.} We visualize how text-image feature alignment evolves across Transformer layers for both image understanding and generation tasks:
(a) Image generation exhibits a rise-then-fall alignment trend across layers.
(b) Image understanding shows an increasing alignment pattern.
(c) When using a fully shared Transformer for both tasks under the next-token prediction objective, the alignment curves converge, reflecting representational compromise between generation and understanding.
(d) Models fine-tuned on Emu3-base~\citep{wang2024emu3} for each individual task recover their distinct trends, consistent with those observed in expert models.}
    \label{fig:sec3:align}
\end{figure}

\textbf{Visual Generation.}
Mainstream visual generative models can be broadly categorized into diffusion-based methods and autoregressive (AR) approaches. Diffusion models~\citep{stablediffusion, sd3, pixart, flux1dev} typically encode images into a continuous latent space and generate them by progressively denoising a sampled Gaussian noise. While these models excel at producing photorealistic images, the discrepancy between their continuous modeling of visual signals and the discrete token-based nature of language generation introduces significant architectural complexity when applied to unified multimodal frameworks. 
In contrast, AR models~\citep{llamagen, dalle, simplear} adopt a GPT-style generation paradigm by discretizing images into token sequences and generating them sequentially. This formulation naturally aligns with the LLMs, making these approachs more suitable for unified multimodal modeling. Representative models such as DALL·E~\citep{dalle} and LlamaGen~\citep{llamagen} exhibit strong instruction-following capabilities and produce high-fidelity images. To further improve generation efficiency while preserving image quality, recent works have introduced alternative generation paradigms, including next-scale generation~\citep{var}, next-neighbor generation~\citep{nextneighbor}, and parallelized generation~\citep{parallegen}.

\textbf{Unified Image Understanding and Generation.}
Unified multimodal models~\citep{wang2024emu3, chameleon} aim to perform both visual understanding and generation within a single architecture, enabling the emergence of more advanced capabilities~\citep{mogao, bagel}. However, previous image generation methods~\citep{sd3, flux1dev} mainly use diffusion-based frameworks with spatial autoencoders, while image understanding methods~\citep{bai2025qwen2.5vl, zhu2025internvl3} typically adopts an AR formulation with semantic encoding. This paradigm gap introduces practical challenges for unifying the two tasks within a single model. To bridge this gap, early approaches~\citep{emu, nextgpt, seedx} introduced external diffusion models for image generation. Other methods directly integrate diffusion objectives into the training of a shared Transformer backbone~\citep{showo, zhou2024transfusion}, removing the need for separate diffusion heads. Representative frameworks such as the Janus series~\citep{wu2025janus, ma2025janusflow, januspro} explicitly decouple visual encoding into dual pathways: a semantic encoder for understanding and a spatial encoder for generation. More recent works introduce task-specific~\citep{wang2025mint, bagel} or modality-specific~\citep{li2025synergen} parameters within the LLM-based Transformer itself, distributing parts of the network to different tasks while maintaining a unified input-output interface.

While these designs improve performance on individual tasks, they often increase architectural and training complexity of the whole framework. Moreover, the relationship between visual understanding and generation remains underexplored, leaving open questions regarding the optimal structure.

\section{Method}

\subsection{Observation and Analysis}\label{section: analysis}
Recent efforts have aimed to unify image understanding and generation within a single framework with various architectures. However, few works have examined the intrinsic relationship between the two tasks. In this study, we investigate their differences through the lens of modality alignment.

Given an image $\mathcal{X}$ and its corresponding textual prompt $\mathcal{T}$, we denote vision features extracted at the $l$-th Transformer layer as $V_l^{\text{gen}} \in \mathbb{R}^{n_v\times c}$ and the textual prompt feature from the final Transformer layer as $T \in \mathbb{R}^{n_t\times c}$. $n_v$ and $n_t$ represent the number of visual and textual tokens respectively, and $c$ is the channel size of the features. 
For the generation task, we sample $500$ prompts from the Geneval~\citep{geneval} dataset. At each layer $l$, we compute the modality alignment score $A_l^{\text{gen}}$ using mutual k-nearest neighbors (mutual-kNN), a commonly used metric for evaluating representation alignment~\citep{platonic}:
\begin{equation*}
    A_l^{\text{gen}} = \text{mutual-kNN}\left(
        \left\{ \frac{1}{n_v} \sum_{i=1}^{n_v} V_{l,i}^{\text{gen}}[b] \right\}_{b=1}^{500},\;
        \left\{ \frac{1}{n_t} \sum_{j=1}^{n_t} T_{j}[b] \right\}_{b=1}^{500}
    \right).
\end{equation*}
For the understanding task, we feed the generated images into the model with the query ``Provide a one-sentence caption for the image:''. We extract the vision feature from each layer $V_l^{\text{und}}$ and compute the alignment score between $V_l^{\text{und}}$ and its corresponding prompt feature in each layer:
\begin{equation*}
    A_l^{\text{und}} = \text{mutual-kNN}\left(
        \left\{ \frac{1}{n_v} \sum_{i=1}^{n_v} V_{l,i}^{\text{und}}[b] \right\}_{b=1}^{500},\;
        \left\{ \frac{1}{n_t} \sum_{j=1}^{n_t} T_{j}[b] \right\}_{b=1}^{500}
    \right).
\end{equation*} 

\textbf{Divergent Alignment Patterns in Generation and Understanding.}
Using this analytical tool, we begin by obtaining the alignment patterns in expert models~\citep{llamagen, llava1.5} trained separately for generation and understanding. As shown in Figure~\ref{fig:sec3:align}(a), we observe that in the generation task, the alignment score increases in early layers but decreases in later layers. This trend is consistent with observations from the REPA~\citep{repa} study on diffusion models, suggesting that early layers focus on cross-modal alignment and semantic grounding, while later layers are responsible for synthesizing high-frequency visual details. In contrast, the understanding task exhibits an increasing alignment score across layers in Figure~\ref{fig:sec3:align}(b), indicating the importance of strong cross-modal alignment in deeper layers for accurate comprehension. These findings reveal that the two tasks require fundamentally different alignment behaviors.

\textbf{Representation Compromise of Fully Shared Backbones under NTP.}
We then examine Emu3-base~\citep{emu}, a native multimodal model pretrained jointly on both tasks. As shown in Figure~\ref{fig:sec3:align}(c), the alignment curves for generation and understanding nearly overlap, both following an increase-then-decrease pattern. This suggests that the understanding task may have compromised the generation objective during training. To validate this, we analyze two task-specific variants finetuned from Emu3-base: Emu3-Gen~\citep{emu} and Emu3-Chat~\citep{emu}. Interestingly, as shown in Figure~\ref{fig:sec3:align}(d), Emu3-Chat recovers the monotonically increasing alignment trend characteristic in understanding tasks, while Emu3-Gen retains the rise-then-fall pattern typical of generation. This further supports our hypothesis that the two tasks prefer different alignment dynamics, and simply sharing a backbone under NTP paradigm may lead to representational conflict.

Motivated by these observations, we propose a Y-shaped architecture that shares early layers for joint semantic learning and decouples later layers to accommodate task-specific alignment needs.


\begin{figure}[!t]
    \centering
    \includegraphics[width=0.9\linewidth]{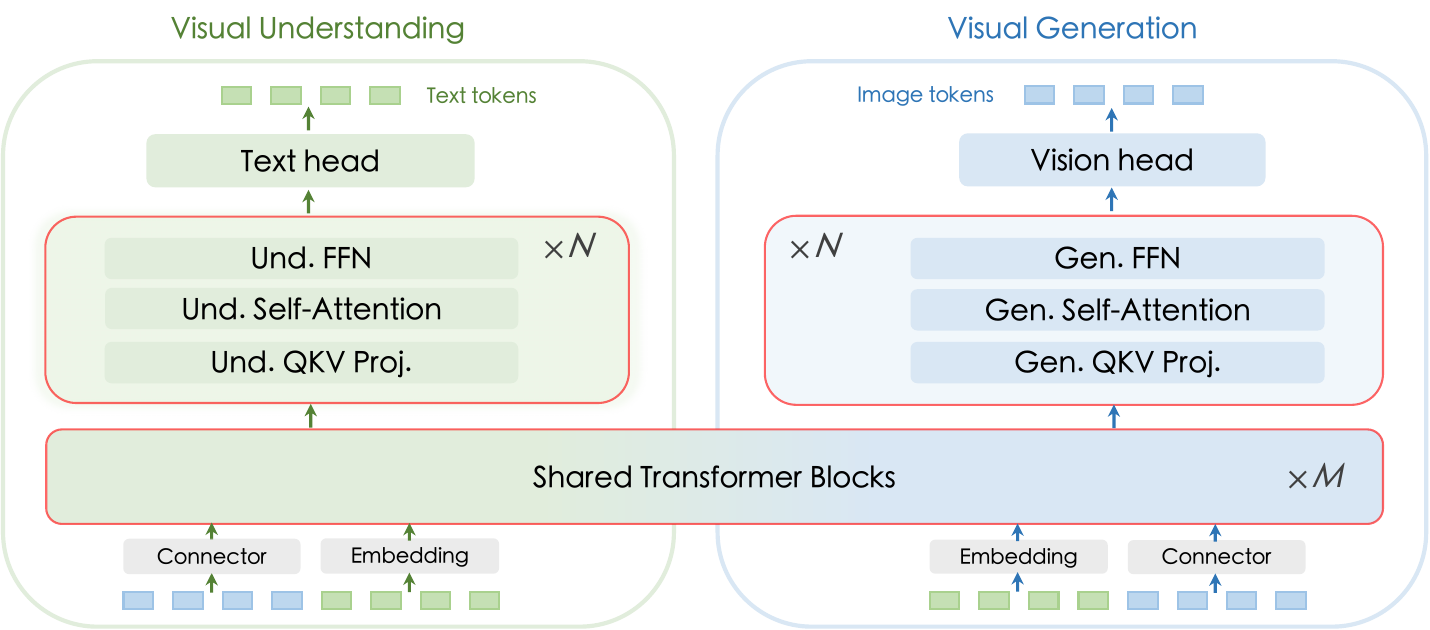}
    \caption{\textbf{Overall framework of UniFork.} UniFork adopts a Y-shaped Transformer backbone. The early layers are shared across both image generation and understanding tasks to facilitate joint semantic representation learning, while the later layers are split into task-specific branches to learn specialized representations. Und.: understanding. Gen.: generation. Proj.: projection.}
    \label{fig:main}
\end{figure}

\subsection{Architecture}
The overall architecture of UniFork is illustrated in Figure~\ref{fig:main}, which enables both learning across tasks and task-specific specialization within a unified framework.

\noindent \textbf{Visual Tokenizer.}
We adopt a single image tokenizer for both understanding and generation to maintain architectural simplicity. Our early exploratory experiments revealed that VAE-based tokenizers perform poorly under limited-scale training, consistent with observations in prior work~\citep{showo}. Instead, we leverage the tokenizer proposed in VILA-U~\citep{vilau}, which preserves image reconstruction quality while enhancing text-image alignment. Given an input image, the tokenizer compresses it by a factor of $16\times16$, flattens the resulting 2D features into a 1D token sequence, and passes it through a lightweight MLP before feeding the tokens into the language model.

\noindent \textbf{Transformer Backbone.}
Motivated by our alignment analysis, UniFork adopts a Y-shaped Transformer architecture. Given a Transformer of $(M+N)$ total layers, the first $M$ layers are shared across both tasks to support joint semantic representation learning. The remaining $N$ layers contains two task-specific branches: one dedicated to semantic reinforcement for image understanding, and the other focusing on spatial detail reconstruction for image generation. We initialize the entire backbone with weights from the  Qwen2.5-0.5B LLM~\citep{qwen2.5}. Notably, when $N = 0$, UniFork reduces to the architecture of Emu3~\citep{wang2024emu3} with full parameter sharing; when $M = 0$, it becomes structurally similar to the recently proposed Mixture-of-Transformers design in BAGEL~\citep{bagel}. 

\noindent \textbf{Generation Vision Head.}
Since the image tokenizer~\citep{vilau} uses the residual vector quantization method ~\citep{rqvae} to map each token into multiple discrete codes, we incorporate an image head to predict these codes. This head takes the output features from the last layer of LLM and generates codes for each token autoregressively.

\subsection{Training Pipeline}
\begin{figure}[!t]
    \centering
    \includegraphics[width=\linewidth]{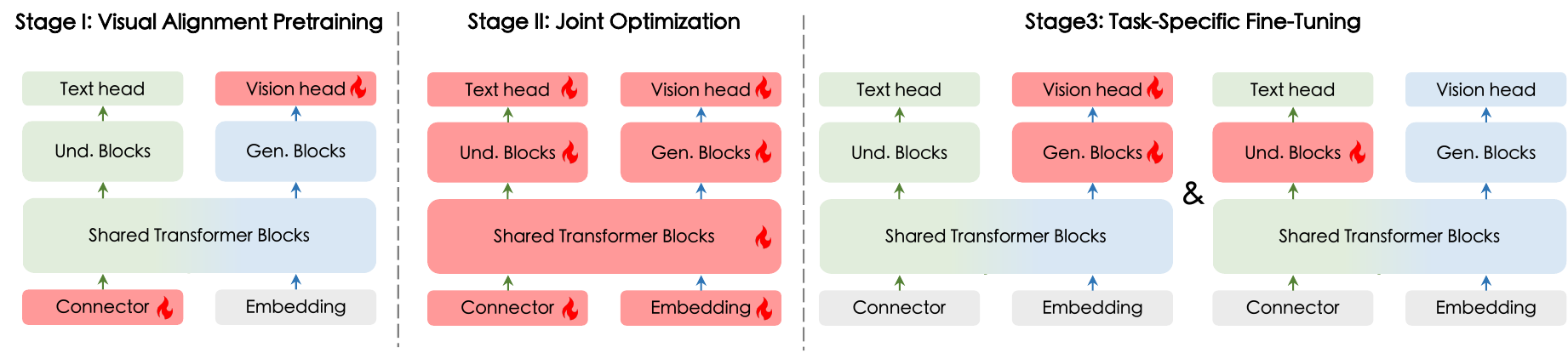}
    \caption{\textbf{Three-stage training pipeline for UniFork.} The first stage focuses on aligning visual and textual modalities. The second stage performs joint training to enhance both image understanding and generation capabilities. In the third stage, task-specific parameters are alternately optimized using data from each task. Modules involved in training are highlighted in \textcolor[rgb]{1.0, 0.369, 0.337}{red}.}
    \label{fig:train}
\end{figure}

As shown in Figure~\ref{fig:train}, the overall training process can be divided into three stages:

\noindent \textbf{Stage I: Visual Alignment Pretraining.} 
The objective of this stage is to align the visual representation with the pretrained LLM. Following prior works~\citep{showo, januspro}, we first train the model on the ImageNet-1K~\citep{imagenet} dataset to efficiently capture pixel-level dependencies. We formulate the learning task using paired images and textual descriptions, where class names are converted into natural language prompts using the OpenAI ImageNet templates~\citep{clip}. This data is used to train both image captioning and text-to-image generation. Subsequently, we perform training on the same two tasks with a mixture of 30 million samples from Laion-En~\citep{laion} and 10 million samples from COYO~\citep{coyo700m}. During this stage, the weights of the LLM are frozen, and we only train the randomly initialized visual connector and image head. The generation task follows the format \texttt{"<caption><image>"}, while the captioning task uses the format \texttt{"<image><caption>"}.

\noindent \textbf{Stage II: Joint Optimization.}
This stage aims to enhance the model's overall ability in both image understanding and generation. We unfreeze the LLM and jointly optimize the backbone, visual connector, and image head. For the multitask pretraining, we use 32.5 million image-text pairs from JourneyDB~\citep{journeydb}, SAM~\citep{sam}, Unsplash~\citep{unsplash}, and an internal dataset for generation, and a 16.5 million subset of InternVL-1.5~\citep{internvl1.5} pretraining data for understanding. We then perform instruction tuning. For generation, we sample a subset from the 32.5 million dataset and combine it with the BLIP3o-60k~\citep{blip3o} dataset, totaling 5 million samples. For understanding, we use a 3.8 million subset of the InternVL-1.5 SFT dataset. The format for generation task is: \texttt{"USER: <Input Message> ASSISTANT: <Response>"}. For the understanding task, we adopt the baseline SFT dialogue format~\citep{qwen2.5}.

\noindent \textbf{Stage III: Task-Specific Fine-Tuning.}
An important advantage of the UniFork architecture is its flexibility in optimization. After joint training, we further refine task-specific performance through isolated fine-tuning. In this stage, only the task-specific layers are updated, while all shared components remain frozen. We reuse the instruction-tuning datasets from Stage II, and independently fine-tune the understanding and generation branches. This final stage allows the model to specialize in each task without introducing interference, effectively balancing shared semantic representation and task-specific optimization.

\subsection{Training Objective}

UniFork models both visual and textual tokens in an autoregressive manner. Therefore, we adopt the cross-entropy loss over both tasks, without introducing any task-specific loss weighting:

\begin{equation}
\mathcal{L}_{\text{total}} = - \sum_{i=1} \log P(\hat{x_i}=x_i \mid x_{<i}),
\end{equation}

where $P$ denotes the probability distribution modeled by the UniFork network. $\hat{x}_i$ and $x_i$ represent the predicted and ground truth token respectively. For the image generation task, the loss is computed only over visual tokens. For the image understanding task, the loss is calculated solely on the response portion of the text tokens.

\section{Experiments}
In this section, we present comprehensive experiments to evaluate the proposed UniFork structure. We begin with the experimental setup (Sec~\ref{sec4: setup}), followed by ablation studies that demonstrate the effectiveness of the Y-shaped architecture (Sec~\ref{sec4: ablation}). Guided by these insights, we modestly scale the model and data, and compare UniFork against both expert models and recent unified models for image understanding and generation (Sec~\ref{sec4: und}, \ref{sec4: gen}). Finally, we analyze the modality alignment patterns of UniFork (Sec~\ref{sec4: align}).

\subsection{Experiment Setup}\label{sec4: setup}

\noindent \textbf{Implementation Details.}
We initialize the Transformer backbone of UniFork using the Qwen2.5-0.5B LLM~\citep{qwen2.5}. The latter half of the Transformer layers are duplicated to construct two independent branches for image understanding and generation, respectively. The total number of parameters in the backbone is 1.21B, with 0.5B active parameters for understanding and 0.76B for generation. We adopt the tokenizer from VILA-U-256~\citep{vilau} to obtain text-aligned codes for each image. The tokenizer has a vocabulary size of $16,384$ with a compression factor of $16\times16$. Input images are resized to a resolution of $384\times384$. For image generation, we resize the shorter side to $384$ and apply center crop on the longer side. For image understanding, the longer side is resized to $384$, and the shorter side is padded with a background color (RGB: $127$, $127$, $127$) to form a $384 \times 384$ input. Following previous works~\citep{llamagen, januspro}, we employ the classifier-free guidance during image generation. Specifically, $10\%$ of input prompts are randomly dropped during training and replaced with a special padding token. During inference, the guidance scale is set to $4.0$ to balance fidelity and diversity. The whole training process is conducted on 16 Nvidia A100 GPUs, detailed configurations for each stage are summarized in Table~\ref{tab:config}.

\begin{table}[!t]
  \centering
  \caption{\textbf{Detailed training configurations of UniFork.}}
    \resizebox{0.8\linewidth}{!}{ 
    \begin{tabular}{c|ccccc}
    \toprule
    \multirow{2}[2]{*}{Configuration} & \multicolumn{1}{c|}{\textbf{Stage 1}} & \multicolumn{2}{c|}{\textbf{Stage 2}} & \multicolumn{2}{c}{\textbf{Stage 3}} \\
          & \multicolumn{1}{r|}{} & Pretrain & \multicolumn{1}{c|}{SFT} & Gen. task & Und. task \\
    \midrule
    Learning rate & \multicolumn{1}{c|}{$1e^{-4}$} & {$5e^{-5}$} & \multicolumn{1}{c|}{{$5e^{-5}$}} & {$3e^{-5}$} & {$2e^{-5}$} \\
    LR scheduler & \multicolumn{1}{c|}{Constant} & Constant & \multicolumn{1}{c|}{Constant} & Constant & Cosine \\
    Warm-up steps & \multicolumn{1}{c|}{0.03} & 0     & \multicolumn{1}{c|}{0} & 0     & 0.03 \\
    Training steps & \multicolumn{1}{c|}{125K} & 170K  & \multicolumn{1}{c|}{23K} & 17K   & 15K \\
    Global batch size & \multicolumn{1}{c|}{384} & 384   & \multicolumn{1}{c|}{384} & 384   & 256 \\
    Weight decay & \multicolumn{5}{c}{0.0} \\
    Optimizer & \multicolumn{5}{c}{AdamW} \\
    Context Length & \multicolumn{5}{c}{$1350$} \\
    Numerical precision & \multicolumn{5}{c}{\texttt{bfloat16}} \\
    \bottomrule
    \end{tabular}}%
  \label{tab:config}%
\end{table}%

\noindent \textbf{Evaluation Benchmarks.}
To evaluate the effectiveness of UniFork in both image understanding and generation, we compare it against expert models trained specifically for each task, as well as recent unified models. For image understanding, we conduct evaluations on five widely adopted benchmarks: MME-P~\citep{mme}, POPE~\citep{pope}, SEED-I~\citep{seedbench}, VQAv2~\citep{vqav2}, and GQA~\citep{gqa}. These benchmarks collectively assess various aspects of visual comprehension, including perception, reasoning, and grounding. For image generation, we use GenEval~\citep{geneval} and MJHQ-30K~\citep{mjhq} benchmarks. GenEval is an object-centric benchmark that assesses text-to-image alignment across six dimensions: ``single objec'', ``two objects'', ``counting'', ``colors'', ``position'', and ``color attributes''. While MJHQ-30K focuses on overall image quality and visual aesthetics. It uses the Fréchet Inception Distance (FID) metric to evaluate the similarity between generated images and a curated set of 30K high-quality reference images.

\subsection{Ablation Study}\label{sec4: ablation}
\noindent \textbf{Effectiveness of UniFork Structure.}
To validate the effectiveness of the proposed UniFork architecture, we conduct a comparative study using four model variants, all initialized from the Qwen2.5-0.5B LLM~\citep{qwen2.5}. The variants include: (1) \textbf{Gen Expert}, trained exclusively on generation data; (2) \textbf{Und Expert}, trained exclusively on understanding data; (3) \textbf{Fully Shared LLM}, which uses a single Transformer backbone for both tasks, with a 0.07B vision head for image generation; and (4) \textbf{UniFork}, which shares the first half layers of the Transformer but adopts separate task-specific branches in the latter half. All models are trained on a subset of the data used in Stage I and Stage II, with the input image resolution set to $256\times256$. To ensure a fair comparison, we keep the number of activated parameters and training configurations consistent for each task.

As shown in Table~\ref{tab:ablation:unifork}, UniFork consistently outperforms the Fully Shared LLM on both image understanding and generation tasks, and achieves comparable or even better performance than the task-specific expert models. These results demonstrate that the Y-shaped Transformer architecture achieves a more effective trade-off between shared semantic learning and task-specific representation. By decoupling the later layers, UniFork reduces task interference and enables targeted feature refinement, leading to overall performance gains across both modalities.

\begin{table}[tbp]
    \vspace{-15pt}
  \centering
  \caption{\textbf{Ablation study to verify the effectiveness of UniFork architecture.} ``Gen Expert'' and ``Und Expert'' dente models trained solely on generation and understanding data, respectively. ``Fully Shared LLM'' denotes a unified model where both tasks share the same Transformer backbone.}
    \resizebox{0.7\linewidth}{!}{ 
    \begin{tabular}{cccc|cc}
    \toprule
    \multirow{2}[4]{*}{Model} & \multicolumn{3}{c|}{Und. Evaluation} & \multicolumn{2}{c}{Gen. Evaluation} \\
\cmidrule{2-6}          & MME-P$\uparrow$ & VQAv2$\uparrow$ & SEED-I$\uparrow$ & Geneval$\uparrow$ & MJHQ$\downarrow$ \\
    \midrule
    Gen Expert & -     & -     & -     & 0.36  & 15.1 \\
    Und Expert & 1177  & 69.7  & 56.6  & -     & - \\
    Fully Shared LLM & 1203  & 69.6  & 53.9  & 0.28  & 17.2 \\
    \cellcolor{rowblue}{UniFork (ablation)} & \cellcolor{rowblue}{1229}  & \cellcolor{rowblue}{69.9} & \cellcolor{rowblue}{55.1}  & \cellcolor{rowblue}{0.33}  & \cellcolor{rowblue}{16.3} \\
    \bottomrule
    \end{tabular}}
  \label{tab:ablation:unifork}%
\end{table}%

\noindent \textbf{Modality Alignment Analysis on More Datasets.}
In Section~\ref{section: analysis}, we analyzed modality alignment patterns using the Geneval~\citep{geneval} benchmark. Its prompts are relatively short and primarily focus on object-level descriptions. To avoid potential dataset-specific biases, we extend our alignment analysis to longer prompts with more emphasis on holistic scene descriptions and stylistic attributes.

\begin{figure}[!t]
    \centering
    \includegraphics[width=\linewidth]{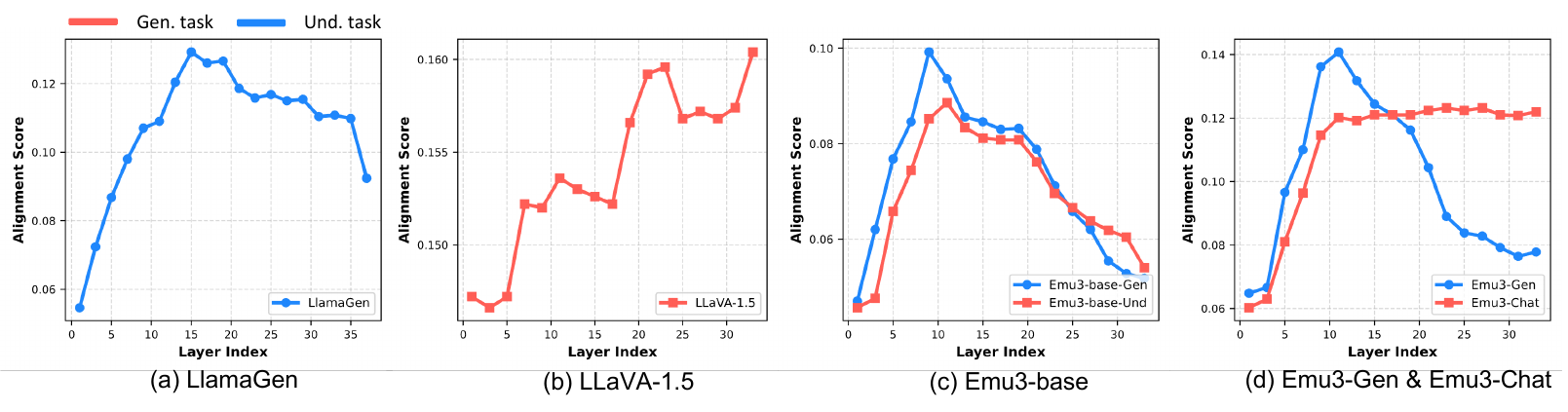}
    \caption{\textbf{Modality alignment analysis on MJHQ-30K.} The observed alignment patterns on this dataset are consistent with those reported in Section~\ref{section: analysis}.}
    \label{fig:alignment-mjhq}
\end{figure}

We randomly sample 500 prompts from MJHQ-30K~\citep{mjhq}, with the average prompt length increasing from $7.3$ words in Geneval~\citep{geneval} to $32.9$ words. As shown in Figure~\ref{fig:alignment-mjhq}, the observed trends remain consistent with those in Figure~\ref{fig:sec3:align}. Specifically, the alignment score for generation still exhibits a rise-then-fall pattern across Transformer layers, while the alignment for understanding continues to increase monotonically. These results further support our earlier findings: fully sharing a Transformer backbone may lead to representational compromise between the two tasks. This highlights the necessity of the UniFork architecture to better accommodate the divergent alignment behaviors of image generation and understanding within a unified framework.

\subsection{Image Understanding Evaluation}\label{sec4: und}
\begin{figure}[htbp]
    \centering
    \includegraphics[width=\linewidth]{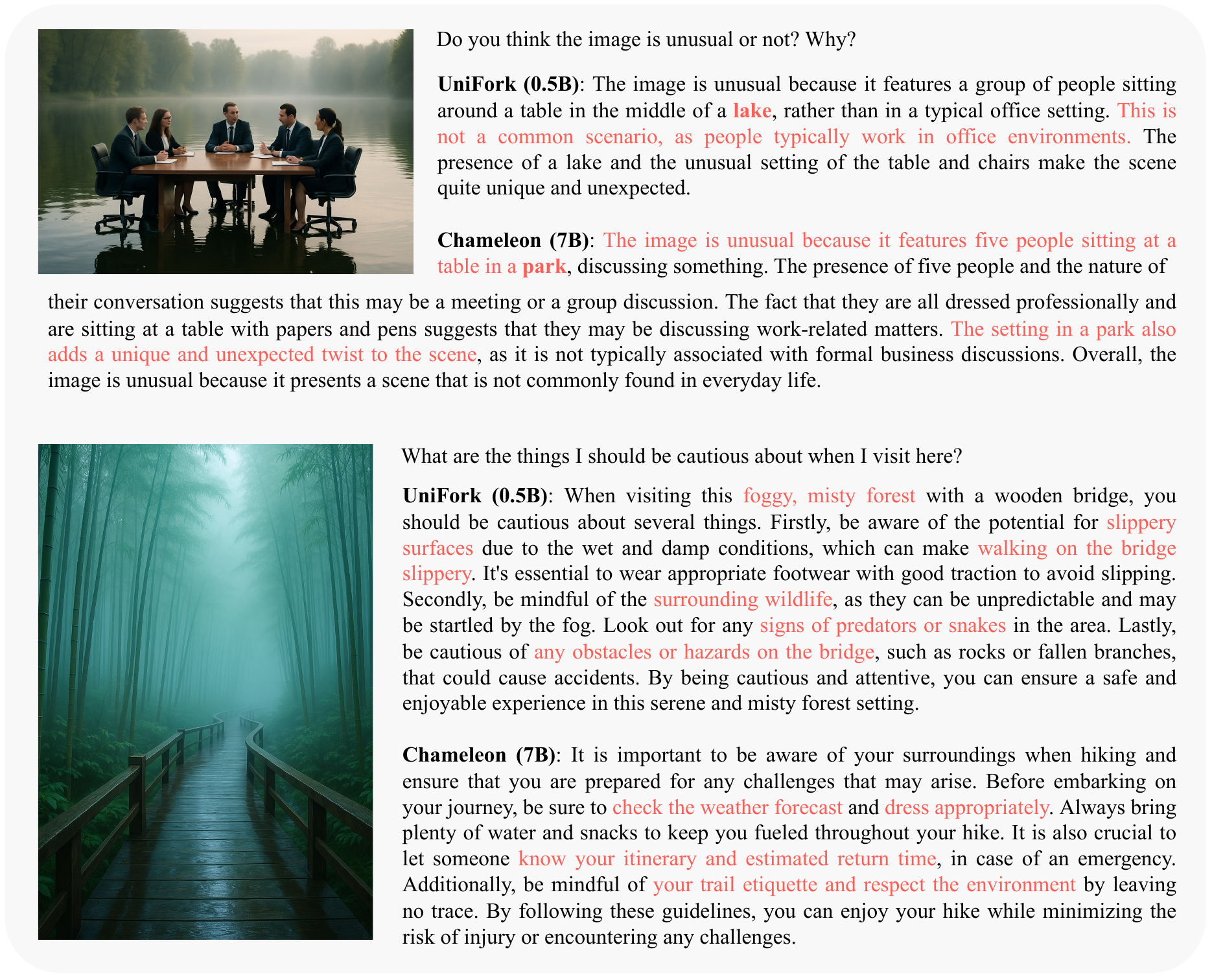}
    \caption{\textbf{Qualitative results on the image understanding task.} The key points in the answers are highlighted in \textcolor[rgb]{1.0, 0.369, 0.337}{red}.}
    \label{fig:quali-und}
\end{figure}

As shown in Table~\ref{tab:und}, UniFork delivers strong performance across all image understanding benchmarks, despite using only 0.5B active parameters for this task. Compared to the recent unified model Show-o~\citep{showo} (1.3B), UniFork achieves relative gains of $10\%$ on MME-P and $7.3\%$ on POPE. Notably, UniFork remains competitive even against larger understanding-only models, such as MobileVLM (2.7B)~\citep{mobilevlm}, IDEFICS-9B~\citep{IDEFICS}, and LLaVA (7B)~\citep{llava}. It matches MobileVLM on POPE ($85.8$ vs. $84.9$), and outperforms IDEFICS-9B on SEEDv1 ($55.2$ vs. $45.0$). We further provide some qualitative results in Figure~\ref{fig:quali-und}. These results highlight the effectiveness of our Y-shaped architecture. It helps reduce task interference and enables UniFork to perform well even with a limited parameter budget.

\begin{table}[htbp]
  \centering
  \caption{\textbf{Evaluation results on multimodal understanding benchmarks.} ``\# LLM A-Params'' denotes the number of activated parameters of Transformer backbone during inference.} 
  \resizebox{\linewidth}{!}{
    \begin{tabular}{clcccccc}
    \toprule
    \textbf{Type} & \textbf{Model} & \textbf{\# LLM A-Params} & \textbf{MME-P}$\uparrow$ & \textbf{POPE}$\uparrow$  & \textbf{SEEDv1}$\uparrow$ & \textbf{VQAv2}$\uparrow$ & \textbf{GQA}$\uparrow$ \\
    \midrule
    \textit{Und. Only} 
          & MobileVLM~\citep{mobilevlm} & 1.4B  & 1196.2 & 84.5  & -     & -     & 56.1 \\
          & MobileVLM~\citep{mobilevlm} & 2.7B  & 1288.9 & 84.9  & -     & -     & 59.0 \\
          & LLaVA~\citep{llava} & 7B    & 809.6 & 76.3  & 33.5  & -     & - \\
          & IDEFICS-9B~\citep{IDEFICS} & 8B    & 1177.3 & 81.9  & 45.0  & -     & 38.4 \\
          & InstructBLIP~\citep{instructblip} & 7B    & -     & -     & 53.4  & -     & 49.5 \\
    & LLaVA-v1.5-Phi-1.5~\citep{showo} & 1.3B  & 1128.0 & 84.1  & -     & 75.3  & - \\
    \midrule
    \textit{Und. and Gen.} & Emu~\citep{emu}   & 13B   & -     & -     & -     & 52.0  & - \\
          & Gemini-Nano-1~\citep{Gemini-Nano} & 1.8B  & -     & -     & -     & 62.7  & - \\
          & LaVIT~\citep{lavit} & 7B    & -     & -     & -     & 66.0  & 46.8 \\
          & LWM~\citep{lwm}   & 7B    & -     & 75.2  & -     & -     & 44.8 \\
          & NExT-GPT~\citep{nextgpt} & 13B   & -     & -     & -     & 66.7  & - \\
          & Chameleon~\citep{chameleon} & 7B    & -     & -     & 30.5  & 66.0  & - \\
          & Show-o~\citep{showo} & 1.3B  & 1097.2 & 80.0  & -     & 69.4  & 58.0 \\
          & D-Dit~\citep{ddit} & 2B    & 1124.7 & 84.0  & -     & -     & 59.2 \\
          \cellcolor{rowblue} & \cellcolor{rowblue}{UniFork (main)} & \cellcolor{rowblue}{0.5B}  & \cellcolor{rowblue}{1208.0} & \cellcolor{rowblue}{85.8} & \cellcolor{rowblue}{55.2} & \cellcolor{rowblue}{70.0} & \cellcolor{rowblue}{55.1} \\
    \bottomrule
    \end{tabular}}%
  \label{tab:und}%
\end{table}%

\subsection{Image Generation Evaluation}\label{sec4: gen}

\begin{figure}[htbp]
    \centering
    \includegraphics[width=\linewidth]{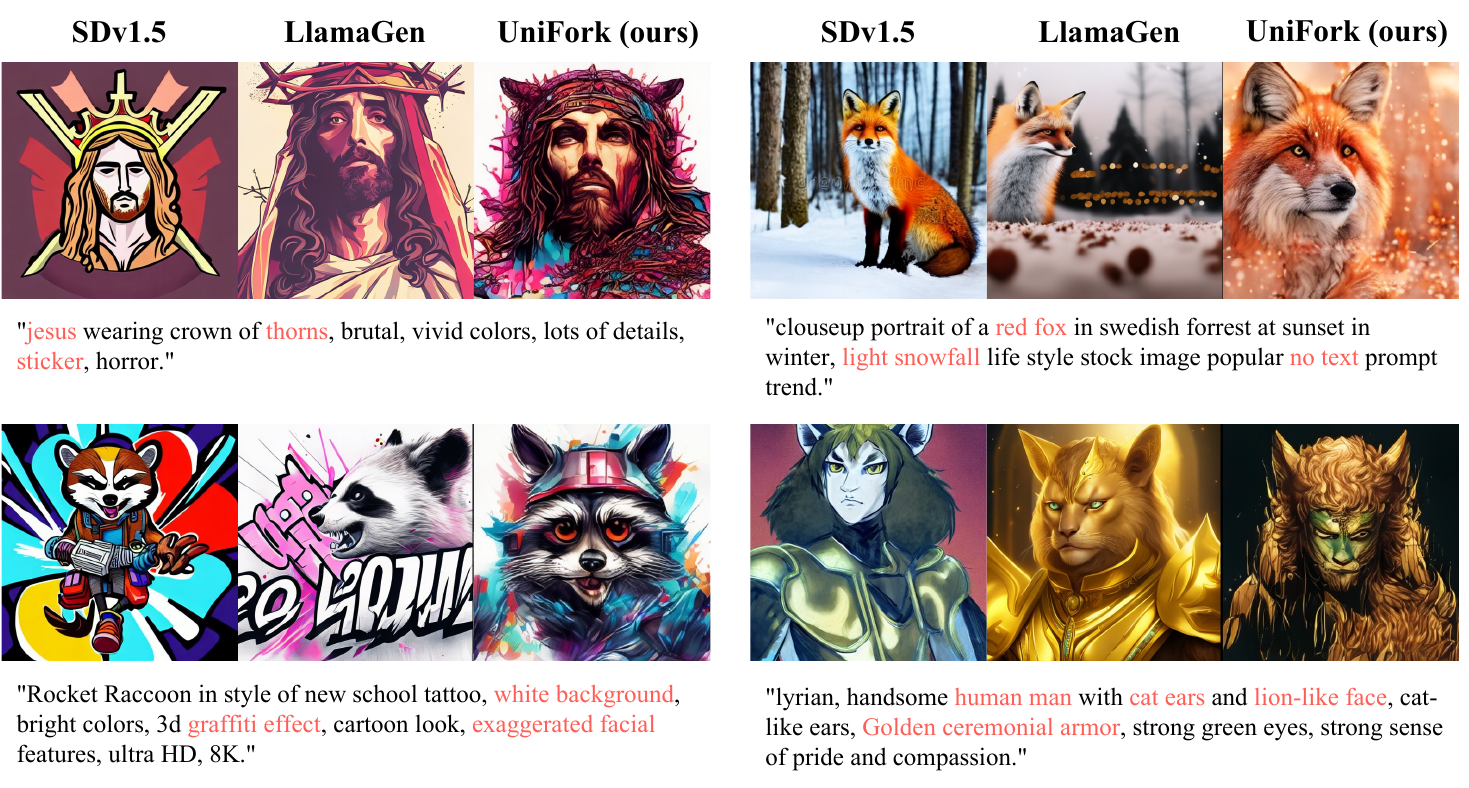}
    \caption{\textbf{Qualitative results on the text-to-image generation task.} We compare image samples generated by SDv1.5, LlamaGen, and UniFork, with respective resolutions of $512\times512$, $512\times512$, and $384\times384$.}
    \label{fig:quali-gen}
\end{figure}

As shown in Table~\ref{tab:geneval}, UniFork achieves an overall $46\%$ accuracy on GenEval, representing a $39\%$ improvement over the ablation variant with smaller parameter scale. Notably, UniFork outperforms prior unified models with similar or larger sizes, such as LWM~\citep{lwm} and Chameleon~\citep{chameleon}, and also surpasses several generation-only baselines, including LDM~\citep{ldm} and LlamaGen~\citep{llamagen}, across most categories. On MJHQ-30K (Table~\ref{tab:mjhq}), UniFork achieves a FID score of $10.6$, marking a $35\%$ improvement over its smaller variant. This FID also surpasses previous unified models such as Show-o ($15.18$)~\citep{showo} and LWM ($17.77$), despite UniFork using significantly fewer parameters (0.76B vs. 7B+). We attribute these gains to the structural insights from our ablation study, which demonstrated that task-specific branches are essential for resolving modality alignment conflicts in unified training. We further provide some qualitative results in Figure~\ref{fig:1} and Figure~\ref{fig:quali-gen}.

By modestly scaling the model from 0.57B to 0.76B active parameters, we unlock substantial performance improvements without requiring architectural changes. We expect further improvements with better tokenizers, larger parameters and higher-quality data.


\begin{table}[htbp]
  \centering
  \caption{\textbf{Image generation results on GenEval dataset.} ``\# LLM A-Params'' denotes the number of activated parameters of Transformer backbone during inference. Obj.: Object. Attri.: Attribution.}
  \resizebox{\linewidth}{!}{
    \begin{tabular}{clccccccc|c}
    \toprule
    \textbf{Type} & \textbf{Model} & \textbf{\# LLM A-Params} & \textbf{Single Obj.} & \textbf{Two Obj.} & \textbf{Counting} & \textbf{Colors} & \textbf{Position} & \textbf{Color Attri.} & \textbf{Overall}$\uparrow$ \\
    \midrule
    \textit{Gen. Only} 
          & LDM~\citep{ldm}   & 1.4B  & 0.92  & 0.29  & 0.23  & 0.70  & 0.02  & 0.05  & 0.37 \\
          & SDv1.5~\citep{ldm} & 0.9B  & 0.97  & 0.38  & 0.35  & 0.76  & 0.04  & 0.06  & 0.43 \\
        & LlamaGen~\citep{llamagen} & 0.8B  & 0.71  & 0.34  & 0.21  & 0.58  & 0.07  & 0.04  & 0.32 \\
    \midrule
          & SEED-X~\citep{seedx} & 17B   & 0.97  & 0.58  & 0.26  & 0.8   & 0.19  & 0.14  & 0.49 \\
          & LWM~\citep{lwm}   & 7B    & 0.93  & 0.41  & 0.46  & 0.79  & 0.09  & 0.15  & 0.47 \\
    \textit{Und. and Gen.} & Chameleon~\citep{chameleon} & 34B   & -     & -     & -     & -     & -     & -     & 0.39 \\
    \cellcolor{rowblue}  &  \cellcolor{rowblue}{UniFork (ablation)} & \cellcolor{rowblue}{0.57B} & \cellcolor{rowblue}{0.83}  & \cellcolor{rowblue}{0.24}  & \cellcolor{rowblue}{0.16}  & \cellcolor{rowblue}{0.62}  & \cellcolor{rowblue}{0.06}  & \cellcolor{rowblue}{0.62}  & \cellcolor{rowblue}{0.33} \\
    \cellcolor{rowblue} &  \cellcolor{rowblue}{UniFork (main)} & \cellcolor{rowblue}{0.76B} & \cellcolor{rowblue}{0.95}  & \cellcolor{rowblue}{0.50}  & \cellcolor{rowblue}{0.22}  & \cellcolor{rowblue}{0.73}  & \cellcolor{rowblue}{0.19}  & \cellcolor{rowblue}{0.17}  & 
          \cellcolor{rowblue}{0.46~(\textcolor{deepred}{↑39\%})}
 \\
    \bottomrule
    \end{tabular}}%
  \label{tab:geneval}%
\end{table}%

\begin{table}[htbp]
  \centering
  \caption{\textbf{Image generation results on MJHQ-30K dataset.} ``\# LLM A-Params'' denotes the number of activated parameters of Transformer backbone during inference.}
  \resizebox{0.6\linewidth}{!}{  
  \begin{tabular}{clcc}
    \toprule
    \textbf{Type}  & \textbf{Model} & \textbf{\# LLM A-Params} & \textbf{MJHQ}$\downarrow$ \\
    \midrule
    \textit{Gen. Only} 
    & LDM~\citep{ldm}      & 1.4B  & - \\
    & SDv1.5~\citep{ldm}   & 0.9B  & - \\
& LlamaGen~\citep{llamagen} & 0.8B  & - \\
    \midrule
    \textit{Und. and Gen.} & LWM~\citep{lwm}             & 7B    & 17.77 \\
        & VILA-U-256~\citep{vilau}   & 7B    & 12.81 \\
        & Show-o~\citep{showo}   & 1.3B  & 15.18 \\
        \cellcolor{rowblue} & \cellcolor{rowblue}{UniFork (ablation)}  & \cellcolor{rowblue}{0.57B} & \cellcolor{rowblue}{16.3} \\
        \cellcolor{rowblue} & \cellcolor{rowblue}{UniFork (main)}  & \cellcolor{rowblue}{0.76B} & 
        \cellcolor{rowblue}{10.6~(\textcolor{deepred}{↑35\%})}
        \\
    \bottomrule
  \end{tabular}}
  \label{tab:mjhq}
\end{table}

\subsection{Modality Alignment Analysis}\label{sec4: align}

\begin{figure}[htbp]
    \centering
    \includegraphics[width=0.4\linewidth]{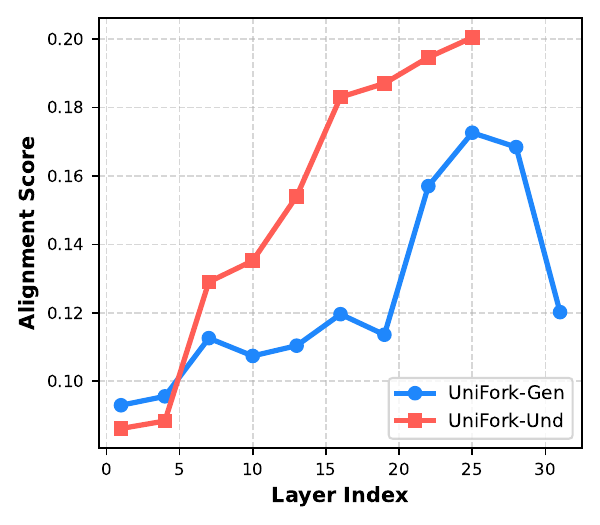}
    \caption{\textbf{Modality alignment analysis for UniFork.}}
    \label{fig:align:unifork}
\end{figure}

Following the methodology introduced in Section~\ref{section: analysis}, we further visualize the modality alignment patterns of UniFork for both image understanding and generation tasks. As shown in Figure~\ref{fig:align:unifork}, the alignment score for the understanding task increases steadily with network depth, while that for the generation task follows a rise-then-fall trend. These trends are consistent with those observed in the expert models, satisfying the distinct representational needs of the two tasks.

This result provides additional evidence for the effectiveness of the UniFork architecture. By decoupling the later layers of the Transformer backbone and assigning task-specific parameters, the model can reconcile the divergent requirements of generation and understanding within a unified framework, without forcing a compromise in representation quality.

\section{Conclusion}
In this paper, we analyzed modality alignment patterns in expert models and NTP-based unified models for image generation and understanding. We found that fully sharing a Transformer backbone may lead to task interference. Inspired by this finding, we proposed UniFork that shares early layers and decouples the later ones for task-specific learning. Ablation studies validate the effectiveness of this design. With modest scaling, UniFork achieves strong performance on both tasks, demonstrating its potential as a baseline for future unified multimodal models.

\noindent \textbf{Limitations.} While UniFork demonstrates strong performance, it remains constrained by three main factors: the quality of the visual tokenizer, the relatively small model size, and the limited quality of the training data. These limitations are particularly pronounced in image generation. In particular, the tokenizer used in UniFork is trained at a resolution of $256$ with a Vision Transformer encoder, whereas the model operates at $384$ resolution, potentially leading to spatial mismatches. Improvement in any of these aspects is likely to yield significant gains in overall performance.

\noindent \textbf{Future Work.}
While UniFork effectively balances shared and task-specific representations, the optimal ratio between these two parts of parameters remains underexplored. This balance likely depends on task complexity, the distribution of training data, and overall model parameters. Future work should also explore scaling up training with interleaved vision-language data to better unlock UniFork’s emergent capabilities, particularly for complex reasoning tasks. Moreover, UniFork’s shared-then-split design provides a flexible foundation for extending beyond vision and language. Incorporating additional modalities, such as audio, video, or 3D data, may offer deeper insights into cross-modal alignment dynamics and support the development of a truly any-to-any multimodal architecture.

\bibliography{iclr2025_conference}
\bibliographystyle{iclr2025_conference}

\end{document}